\documentclass[runningheads]{llncs}
\usepackage{graphicx}
\usepackage{amsmath,amssymb} 
\usepackage{color}
\usepackage[width=122mm,left=12mm,paperwidth=146mm,height=193mm,top=12mm,paperheight=217mm]{geometry}
\usepackage{multirow}
\usepackage{booktabs}
\usepackage{bm}

\usepackage{cite}
\usepackage{makecell}
\usepackage{boldline}
\setcellgapes{3pt}
\usepackage{tabu}
\usepackage{subcaption}
\usepackage[font=small,labelfont=bf]{caption}
\usepackage[pagebackref=true,breaklinks=true,letterpaper=true,colorlinks,bookmarks=false]{hyperref}

\begin{document}
\pagestyle{headings}
\mainmatter

\title{A Trilateral Weighted Sparse Coding Scheme \\ for Real-World Image Denoising}

\titlerunning{Trilateral Weighted Sparse Coding Scheme for Real-World Image Denoising}

\authorrunning{J. Xu, L. Zhang, and D. Zhang}

\author{Jun Xu$^{1}$, Lei Zhang$^{1}$\thanks{This project is supported by Hong Kong RGC GRF project (PolyU 152124/15E).}, David Zhang$^{1, 2}$}


\institute{$^{1}$The Hong Kong Polytechnic University, Hong Kong SAR, China\\
$^{2}$School of Science and Engineering, The Chinese University of Hong Kong (Shenzhen), Shenzhen, China\\
	\email{ \{csjunxu, cslzhang, csdzhang\}@comp.polyu.edu.hk}
}

\maketitle

\begin{abstract}
Most of existing image denoising methods assume the corrupted noise to be additive white Gaussian noise (AWGN). However, the realistic noise in real-world noisy images is much more complex than AWGN, and is hard to be modeled by simple analytical distributions. As a result, many state-of-the-art denoising methods in literature become much less effective when applied to real-world noisy images captured by CCD or CMOS cameras. In this paper, we develop a trilateral weighted sparse coding (TWSC) scheme for robust real-world image denoising. Specifically, we introduce three weight matrices into the data and regularization terms of the sparse coding framework to characterize the statistics of realistic noise and image priors. TWSC can be reformulated as a linear equality-constrained problem and can be solved by the alternating direction method of multipliers. The existence and uniqueness of the solution and convergence of the proposed algorithm are analyzed. Extensive experiments demonstrate that the proposed TWSC scheme outperforms state-of-the-art denoising methods on removing realistic noise.
\keywords{real-world image denoising, sparse coding}
\end{abstract}

\section{Introduction}
Noise will be inevitably introduced in imaging systems and may severely damage the quality of acquired images.\ Removing noise from the acquired image is an essential step in photography and various computer vision tasks such as segmentation \cite{zhu2017non}, HDR imaging \cite{autonoisemodel}, and recognition \cite{nguyen2015deep}, etc.\ Image denoising aims to recover the clean image $\mathbf{x}$ from its noisy observation $\mathbf{y}=\mathbf{x}+\mathbf{n}$, where $\mathbf{n}$ is the corrupted noise.\ This problem has been extensively studied in literature, and numerous statistical image modeling and learning methods have been proposed in the past decades \cite{ksvd,srcolor,nlm,bm3d,cbm3d,bm3dsapca,lssc,ncsr,saist,wnnm,pgpd,foe,epll,mlp,csf,chen2015learning,
dncnn,Liu2008,noiseclinic,Zhu_2016_CVPR,crosschannel2016,neatimage,mcwnnm}. 

\begin{figure}[t!]
\centering
\vspace{-1mm}
\begin{subfigure}[t]{0.24\textwidth}
\raisebox{-\height}{\includegraphics[width=\textwidth]{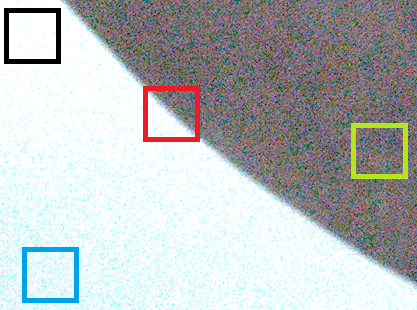}}
\end{subfigure}
\begin{subfigure}[t]{0.24\textwidth}
\raisebox{-\height}{\includegraphics[width=0.37\textwidth]{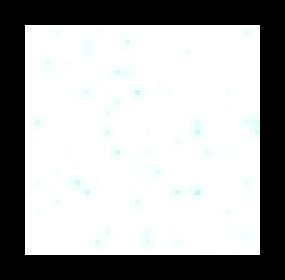}}
\raisebox{-\height}{\includegraphics[width=0.37\textwidth]{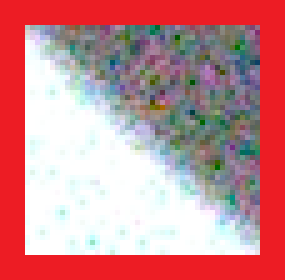}}
\\
\raisebox{-\height}{\includegraphics[width=0.37\textwidth]{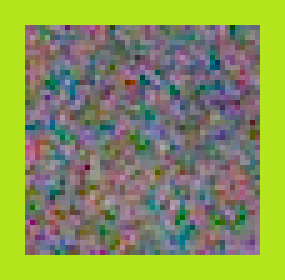}}
\raisebox{-\height}{\includegraphics[width=0.37\textwidth]{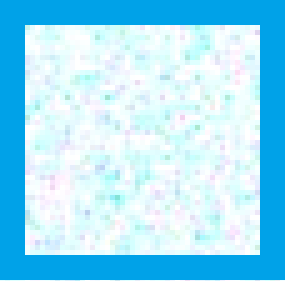}}
\end{subfigure}
\begin{subfigure}[t]{0.24\textwidth}
\raisebox{-\height}{\includegraphics[width=\textwidth]{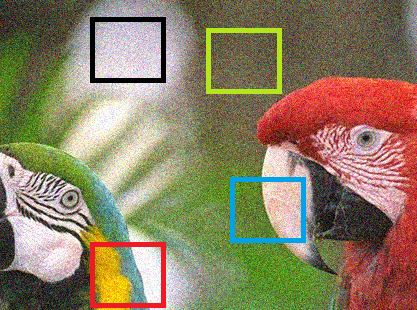}}
\end{subfigure}
\begin{subfigure}[t]{0.24\textwidth}
\raisebox{-\height}{\includegraphics[width=0.42\textwidth]{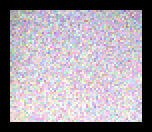}}
\raisebox{-\height}{\includegraphics[width=0.42\textwidth]{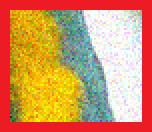}}%
\\
\raisebox{-\height}{\includegraphics[width=0.42\textwidth]{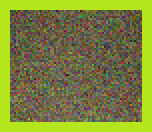}}
\raisebox{-\height}{\includegraphics[width=0.42\textwidth]{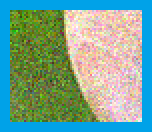}}
\end{subfigure}
\vspace{-3mm}
\caption{\small Comparison of noisy image patches in real-world noisy image (left) and synthetic noisy image with additive white Gaussian noise (right).}
\vspace{-2mm}
\label{f1}
\end{figure}

\begin{figure}[t!]
\vspace{-2mm}
\centering
\begin{subfigure}[t]{0.155\textwidth}
\raisebox{-0.15cm}{\includegraphics[width=1\textwidth]{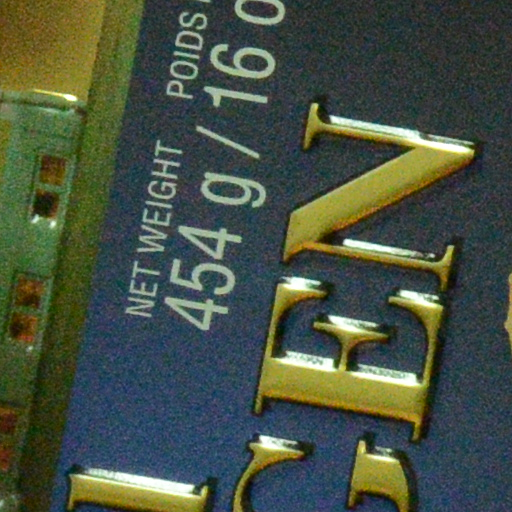}}
\centering{\scriptsize (a)}
\end{subfigure}
\begin{subfigure}[t]{0.155\textwidth}
\raisebox{-0.15cm}{\includegraphics[width=1\textwidth]{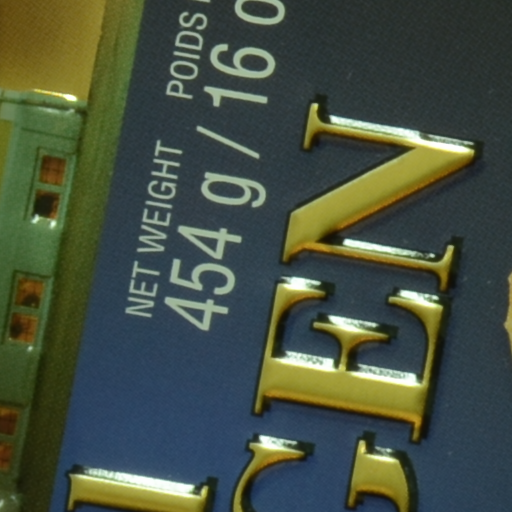}}
\centering{\scriptsize (b)}
\end{subfigure}
\begin{subfigure}[t]{0.155\textwidth}
\raisebox{-0.15cm}{\includegraphics[width=1\textwidth]{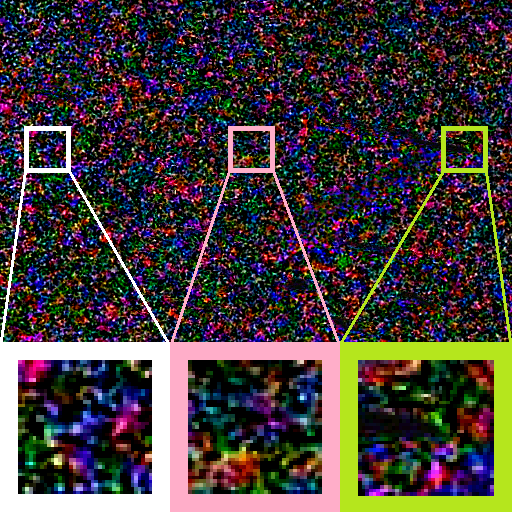}}
\centering{\scriptsize (c)}
\end{subfigure}
\begin{subfigure}[t]{0.155\textwidth}
\raisebox{-0.15cm}{\includegraphics[width=1\textwidth]{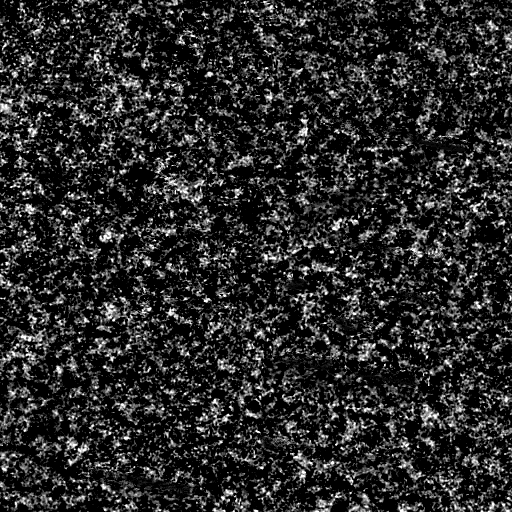}}
\centering{\scriptsize (d)}
\end{subfigure}
\begin{subfigure}[t]{0.155\textwidth}
\raisebox{-0.15cm}{\includegraphics[width=1\textwidth]{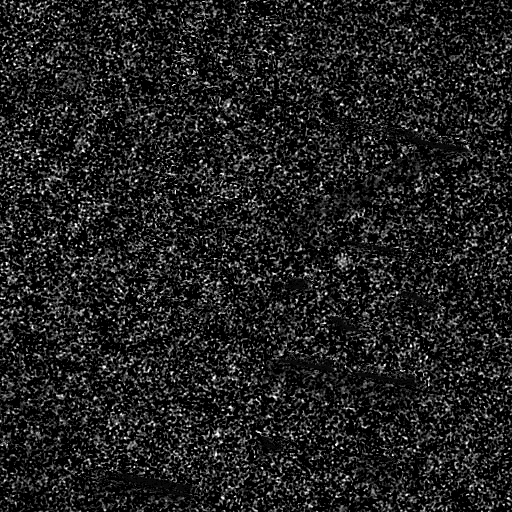}}
\centering{\scriptsize (e)}
\end{subfigure}
\begin{subfigure}[t]{0.155\textwidth}
\raisebox{-0.15cm}{\includegraphics[width=1\textwidth]{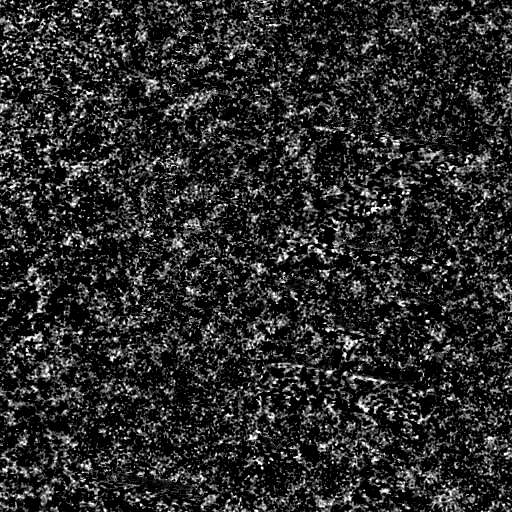}}
\centering{\scriptsize (f)}
\end{subfigure}
\vspace{-3mm}
\caption{\small An example of realistic noise.\ (a) A real-world noisy image captured by a Nikon D800 camera with $\text{ISO}=6400$; (b) the ``Ground Truth'' image (please refer to Section 4.3) of (a); (c) difference between (a) and (b) (amplified for better illustration); (d)-(f) red, green, and blue channel of (c), respectively.\ The standard deviations (stds) of noise in the three boxes (white, pink, and green) plotted in (c) are 5.2, 6.5, and 3.3, respectively, while the stds of noise in each channel (d), (e), and (f) are 5.8, 4.4, and 5.5, respectively.}
\label{f2}
\vspace{-6mm}
\end{figure}

Most of the existing methods \cite{ksvd,srcolor,nlm,bm3d,cbm3d,bm3dsapca,lssc,ncsr,saist,wnnm,foe,epll,pgpd,mlp,csf,chen2015learning,dncnn} focus on additive white Gaussian noise (AWGN), and they can be categorized into dictionary learning based methods \cite{ksvd,srcolor}, nonlocal self-similarity based methods \cite{nlm,bm3d,cbm3d,bm3dsapca,lssc,ncsr,saist,wnnm,pgpd}, sparsity based methods \cite{ksvd,srcolor,bm3d,cbm3d,bm3dsapca,lssc,ncsr}, low-rankness based methods \cite{saist,wnnm}, generative learning based methods \cite{foe,epll,pgpd}, and discriminative learning based methods \cite{mlp,csf,chen2015learning,dncnn}, etc.\ However, the realistic noise in real-world images captured by CCD or CMOS cameras is much more complex than AWGN \cite{Liu2008,noiseclinic,Zhu_2016_CVPR,crosschannel2016,xuaccv2016,mcwnnm,gid2018}, which can be signal dependent and vary with different cameras and camera settings (such as ISO, shutter speed, and aperture, etc.).\ In Fig.\ \ref{f1}, we show a real-world noisy image from the Darmstadt Noise Dataset (DND) \cite{dnd2017} and a synthetic AWGN image from the Kodak PhotoCD Dataset (\url{http://r0k.us/graphics/kodak/}).\ We can see that the different local patches in real-world noisy image show different noise statistics, e.g., the patches in black and blue boxes show different noise levels although they are from the same white object.\ In contrast, all the patches from the synthetic AWGN image show homogeneous noise patterns.\ Besides, the realistic noise varies in different channels as well as different local patches \cite{noiseclinic,Zhu_2016_CVPR,crosschannel2016,mcwnnm}.\ In Fig.\ \ref{f2}, we show a real-world noisy image captured by a Nikon D800 camera with ISO=6400, its ``Ground Truth'' (please refer to Section 4.3), and their differences in full color image as well as in each channel.\ The overall noise standard deviations (stds) in Red, Green, and Blue channels are 5.8, 4.4, and 5.5, respectively.\ Besides, the realistic noise is inhomogeneous.\ For example, the stds of noise in the three boxes plotted in Fig.\  \ref{f2} (c) vary largely.\ Indeed, the noise in real-world noisy image is much more complex than AWGN noise.\ Though having shown promising performance on AWGN noise removal, many of the above mentioned methods \cite{ksvd,srcolor,nlm,bm3d,cbm3d,bm3dsapca,lssc,ncsr,saist,wnnm,foe,epll,pgpd,mlp,
csf,chen2015learning,dncnn} will become much less effective when dealing with the complex realistic noise as shown in Fig.\ \ref{f2}.

In the past decade, several denoising methods for real-world noisy images have been developed \cite{Liu2008,noiseclinic,Zhu_2016_CVPR,
crosschannel2016,mcwnnm,neatimage}.\ Liu et al. \cite{Liu2008} proposed to estimate the noise via a ``noise level function'' and remove the noise for each channel of the real image.\ However, processing each channel separately would often achieve unsatisfactory performance and generate artifacts \cite{srcolor}.\ The methods \cite{noiseclinic,Zhu_2016_CVPR} perform image denoising by concatenating the patches of RGB channels into a vector.\ However, the concatenation does not consider the different noise statistics among different channels.\ Besides, the method of \cite{Zhu_2016_CVPR} models complex noise via mixture of Gaussian distribution, which is time-consuming due to the use of variational Bayesian inference techniques.\ The method of \cite{crosschannel2016} models the noise in a noisy image by a multivariate Gaussian and performs denoising by the Bayesian non-local means \cite{kervrann2007bayesian}.\ The commercial software Neat Image \cite{neatimage} estimates the global noise parameters from a flat region of the given noisy image and filters the noise accordingly.\ However, both the two methods \cite{crosschannel2016,neatimage} ignore the local statistical property of the noise which is signal dependent and varies with different pixels.\ The method \cite{mcwnnm} considers the different noise statistics in different channels, but ignores that the noise is signal dependent and has different levels in different local patches.\ By far, real-world image denoising is still a challenging problem in low level vision \cite{dnd2017}.  

Sparse coding (SC) has been well studied in many computer vision and pattern recognition problems \cite{wright2009robust,yang2009linear,yang2010image}, including image denoising \cite{ksvd,srcolor,lssc,ncsr,pgpd}.\ In general, given an input signal $\mathbf{y}$ and the dictionary $\mathbf{D}$ of coding atoms, the SC model can be formulated as
\vspace{-5mm}
\begin{equation}
\vspace{-5mm}
\label{e1}
\min_{\mathbf{c}}
\|
\mathbf{y}-\mathbf{D}\mathbf{c}
\|_{2}^{2}
+
\lambda\|\mathbf{c}\|_{q},
\end{equation}
where $\mathbf{c}$ is the coding vector of the signal $\mathbf{y}$ over the dictionary $\mathbf{D}$, $\lambda$ is the regularization parameter, and $q=0$ or $1$ to enforce sparse regularization on $\mathbf{c}$.\ Some representative SC based image denoising methods include K-SVD \cite{ksvd}, LSSC \cite{lssc}, and NCSR \cite{ncsr}.\ Though being effective on dealing with AWGN, SC based denoising methods are essentially limited by the data-fidelity term described by $\ell_{2}$ (or Frobenius) norm, which actually assumes white Gaussian noise and is not able to characterize the signal dependent and realistic noise.

In this paper, we propose to lift the SC model (\ref{e1}) to a robust denoiser for real-world noisy images by utilizing the channel-wise statistics and locally signal dependent property of the realistic noise, as demonstrated in Fig.\  \ref{f2}.\ Specifically, we propose a trilateral weighted sparse coding (TWSC) scheme for real-world image denoising.\ Two weight matrices are introduced into the data-fidelity term of the SC model to characterize the realistic noise property, and another weight matrix is introduced into the regularization term to characterize the sparsity priors of natural images.\ We reformulate the proposed TWSC scheme into a linear equality-constrained optimization program, and solve it under the alternating direction method of multipliers (ADMM) \cite{admm} framework.\ One step of our ADMM is to solve a Sylvester equation, whose unique solution is not always guaranteed.\ Hence, we provide theoretical analysis on the existence and uniqueness of the solution to the proposed TWSC scheme.\ Experiments on three datasets of real-world noisy images demonstrate that the proposed TWSC scheme achieves much better performance than the state-of-the-art denoising methods.

\section{The Proposed Real-World Image Denoising Algorithm}

\vspace{-2mm}
\subsection{The Trilateral Weighted Sparse Coding Model}
\vspace{-2mm}

The real-world image denoising problem is to recover the clean image from its noisy observation.\ Current denoising methods \cite{ksvd,srcolor,nlm,bm3d,cbm3d,bm3dsapca,lssc,ncsr,saist,wnnm,pgpd,foe,epll} are mostly patch based. Given a noisy image, a local patch of size $p\times p \times 3$ is extracted from it and stretched to a vector, denoted by $\mathbf{y}=[\mathbf{y}_{r}^{\top}\ \mathbf{y}_{g}^{\top}\ \mathbf{y}_{b}^{\top}]^{\top}\in\mathbb{R}^{3p^{2}}$, where $\mathbf{y}_{c}\in\mathbb{R}^{p^{2}}$ is the corresponding patch in channel $c$, where $c\in\{r,g,b\}$ is the index of R, G, and B channels.\ For each local patch $\mathbf{y}$, we search the $M$ most similar patches to it (including $\mathbf{y}$ itself) by Euclidean distance in a local window around it.\ By stacking the $M$ similar patches column by column, we form a noisy patch matrix $\mathbf{Y}=\mathbf{X}+\mathbf{N}\in\mathbb{R}^{3p^{2}\times M}$, where $\mathbf{X}$ and $\mathbf{N}$ are the corresponding clean and noise patch matrices, respectively.\ The noisy patch matrix can be written as $\mathbf{Y}=[\mathbf{Y}_{r}^{\top}\ \mathbf{Y}_{g}^{\top}\ \mathbf{Y}_{b}^{\top}]^{\top}$, where $\mathbf{Y}_{c}$ is the sub-matrix of channel $c$.\ Suppose that we have a dictionary $\mathbf{D}=[\mathbf{D}_{r}^{\top}\ \mathbf{D}_{g}^{\top}\ \mathbf{D}_{b}^{\top}]^{\top}$, where $\mathbf{D}_{c}$ is the sub-dictionary corresponding to channel $c$.\ In fact, the dictionary $\mathbf{D}$ can be learned from external natrual images, or from the input noisy patch matrix $\mathbf{Y}$. 

Under the traditional sparse coding (SC) framework \cite{lasso}, the sparse coding matrix of $\mathbf{Y}$ over $\mathbf{D}$ can be obtained by
\vspace{-4mm}
\begin{equation}
\vspace{-4mm}
\label{e2}
\hat{\mathbf{C}}
=
\arg\min_{\mathbf{C}}
\|\mathbf{Y}-\mathbf{D}\mathbf{C}\|_{F}^{2}
+
\lambda\|\mathbf{C}\|_{1},
\end{equation}
where $\lambda$ is the regularization parameter.\ Once $\hat{\mathbf{C}}$ is computed, the latent clean patch matrix $\hat{\mathbf{X}}$ can be estimated as $\hat{\mathbf{X}}=\mathbf{D}\hat{\mathbf{C}}$.\ Though having achieved promising performance on additive white Gaussian noise (AWGN), the tranditional SC based denoising methods \cite{ksvd,srcolor,lssc,ncsr,pgpd} are very limited in dealing with realistic noise in real-world images captured by CCD or CMOS cameras.\ The reason is that the realistic noise is non-Gaussian, varies locally and across channels, which cannot be characterized well by the Frobenius norm in the SC model (\ref{e2}) \cite{Liu2008,crosschannel2016,dnd2017,jddcrf}. 

To account for the varying statistics of realistic noise in different channels and different patches, we introduce two weight matrices $\mathbf{W}_{1}\in\mathbb{R}^{3p^2\times3p^2}$ and $\mathbf{W}_{2}\in\mathbb{R}^{M\times M}$ to characterize the SC residual ($\mathbf{Y}-\mathbf{D}\mathbf{C}$) in the data-fidelity term of Eq.\ (\ref{e2}).\ Besides, to better characterize the sparsity priors of the natural images, we introduce a third weight matrix $\mathbf{W}_{3}$, which is related to the distribution of the sparse coefficients matrix $\mathbf{C}$, into the regularization term of Eq.\ (\ref{e2}).\ For the dictionary $\mathbf{D}$, we learn it adaptively by applying the SVD \cite{svd} to the given data matrix $\mathbf{Y}$ as 
\vspace{-3mm}
\begin{equation}
\vspace{-3mm}
\label{e3}
\mathbf{Y} =\mathbf{D}\mathbf{S}\mathbf{V}^{\top}.
\end{equation}
Note that in this paper, we are not aiming at proposing a new dictionary learning scheme as \cite{ksvd} did.\ Once obtained from SVD, the dictionary $\mathbf{D}$ is fixed and not updated iteratively.\ Finally, the proposed trilateral weighted sparse coding (TWSC) model is formulated as:
\vspace{-4mm}
\begin{equation}
\vspace{-4mm}
\label{e4}
\min_{\mathbf{C}}\|\mathbf{W}_{1}(\mathbf{Y}-\mathbf{D}\mathbf{C})\mathbf{W}_{2}\|_{F}^{2}
+
\|\mathbf{W}_{3}^{-1}\mathbf{C}\|_{1}.
\end{equation}
Note that the parameter $\lambda$ has been implicitly incorporated into the weight matrix  $\mathbf{W}_{3}$. 

\vspace{-2mm}
\subsection{The Setting of Weight Matrices}
\vspace{-1mm}

In this paper, we set the three weight matrices $\mathbf{W}_{1}$, $\mathbf{W}_{2}$, and $\mathbf{W}_{3}$ as diagonal matrices and grant clear physical meanings to them.\ $\mathbf{W}_{1}$ is a block diagonal matrix with three blocks, each of which has the same diagonal elements to describe the noise properties in the corresponding R, G, or B channel.\ Based on \cite{Leungtip,dnd2017,jddcrf}, the realistic noise in a local patch could be approximately modeled as Gaussian, and each diagonal element of $\mathbf{W}_{2}$ is used to describe the noise variance in the corresponding patch $\mathbf{y}$.\ Generally speaking, $\mathbf{W}_{1}$ is employed to regularize the row discrepancy of residual matrix ($\mathbf{Y}-\mathbf{D}\mathbf{C}$), while $\mathbf{W}_{2}$ is employed to regularize the column discrepancy of ($\mathbf{Y}-\mathbf{D}\mathbf{C}$).\ For matrix $\mathbf{W}_{3}$, each diagonal element is set based on the sparsity priors on $\mathbf{C}$.

We determine the three weight matrices $\mathbf{W}_{1}$, $\mathbf{W}_{2}$, and $\mathbf{W}_{3}$ by employing the \textsl{Maximum A-Posterior} (MAP) estimation technique:
\vspace{-4mm}
\begin{equation}
\vspace{-4mm}
\begin{split}
\label{e5}
\hat{\mathbf{C}} 
&
=
\arg\max_{\mathbf{C}}\ln P(\mathbf{C}|\mathbf{Y})
=
\arg\max_{\mathbf{C}}\{\ln P(\mathbf{Y}|\mathbf{C})+\ln P(\mathbf{C})\}.
\end{split}
\end{equation}
The log-likelihood term $\ln P(\mathbf{Y}|\mathbf{C})$ is characterized by the
statistics of noise.\ According to \cite{Leungtip,dnd2017,jddcrf}, it can be assumed that the noise is independently and identically distributed (i.i.d.) in each channel and each patch with Gaussian distribution.\ Denote by $\mathbf{y}_{cm}$ and $\mathbf{c}_{m}$ the $m$th column of the matrices $\mathbf{Y}_{c}$ and $\mathbf{C}$, respectively, and denote by $\sigma_{cm}$ the noise std of $\mathbf{y}_{cm}$. We have
\vspace{-4mm}
\begin{equation}
\vspace{-4mm}
\begin{split}
\label{e6}
\hspace{-3mm}
P(\mathbf{Y}|\mathbf{C}) 
&
= 
\hspace{-3mm}
\prod_{c\in\{r, g, b\}}
\prod_{m=1}^{M}
(\pi\sigma_{cm})^{-p^{2}}
e^{-\sigma_{cm}^{-2}
\|
\mathbf{y}_{cm}
-
\mathbf{D}_{c}\mathbf{c}_{m}
\|_{2}^{2}}.
\end{split}
\end{equation}
From the perspective of statistics \cite{glm}, the set of $\{\sigma_{cm}\}$ can be viewed as a $3\times M$ contingency table created by two variables $\sigma_{c}$ and $\sigma_{m}$, and their relationship could be modeled by a log-linear model $\sigma_{cm}=\sigma_{c}^{l_{1}}\sigma_{m}^{l_{2}}$, where $l_{1}+l_{2}=1$.\ Here we consider $\{\sigma_{c}\}, \{\sigma_{m}\}$ of equal importance and empirically set $l_{1}=l_{2}=1/2$.\ The estimation of $\{\sigma_{cm}\}$ can be transferred to the estimation of $\{\sigma_{c}\}$ and $\{\sigma_{m}\}$, which will be introduced in the experimental section (Section 4).

The sparsity prior is imposed on the coefficients matrix $\mathbf{C}$, we assume that each column $\mathbf{c}_{m}$ of $\mathbf{C}$ follows i.i.d. Laplacian distribution.\ Specifically, for each entry $\mathbf{c}_{m}^{i}$, which is the coding coefficient of the $m$th patch $\mathbf{y}_{m}$ over the $i$th atom of dictionary $\mathbf{D}$, we assume that it follows distribution of $(2\bm{S}_{i})^{-1}\exp(-\mathbf{S}_{i}^{-1}|\mathbf{c}_{m}^{i}|)$, where $\mathbf{S}_{i}$ is the $i$th diagonal element of the singular value matrix $\mathbf{S}$ in Eq.\ (\ref{e3}).\ Note that we set the scale factor of the distribution as the inverse of the $i$th singular value $\mathbf{S}_{i}$.\ This is because the larger the singular value $\mathbf{S}_{i}$ is, the more important the $i$th atom (i.e., singular vector) in $\mathbf{D}$ should be, and hence the distribution of the coding coefficients over this singular vector should have stronger regularization with weaker sparsity.\ The prior term in Eq.\ (\ref{e5}) becomes
\vspace{-4mm}
\begin{equation}
\vspace{-4mm}
\label{e7}
P(\mathbf{C})
=
\prod_{m=1}^{M}
\prod_{i=1}^{3p^{2}}
(2\mathbf{S}_{i})^{-1}e^{-\mathbf{S}_{i}^{-1}|\mathbf{c}_{m}^{i}|}.
\end{equation}

Put (\ref{e7}) and (\ref{e6}) into (\ref{e5}) and consider the log-linear model $\sigma_{cm}=\sigma_{c}^{1/2}\sigma_{m}^{1/2}$, we have
\vspace{-4mm}
\begin{equation}
\vspace{-4mm}
\begin{split}
\label{e8}
\hspace{-0mm}
\hat{\mathbf{C}}
&
=
\arg\min_{\mathbf{C}}
\hspace{-4mm}
\sum_{c\in\{r, g, b\}}
\hspace{-1mm}
\sum_{m=1}^{M}
\hspace{-1mm}
\sigma_{cm}^{-2}
\|\mathbf{y}_{cm}
\hspace{-1mm}
-
\hspace{-1mm}
\mathbf{D}_{c}\mathbf{c}_{m}\|_{2}^{2}
\hspace{-0mm}
+
\hspace{-1mm}
\sum_{m=1}^{M}
\hspace{-1mm}
\|\mathbf{S}^{-1}\mathbf{c}_{m}\|_{1}
\\
&
=
\arg\min_{\mathbf{C}}
\hspace{-4mm}
\sum_{c\in\{r, g, b\}}
\hspace{-3mm}
\sigma_{c}^{-1}
\|(\mathbf{Y}_{c}-\mathbf{D}_{c}\mathbf{C})\mathbf{W}_{2}\|_{F}^{2}+\|\mathbf{S}^{-1}\mathbf{C}\|_{1}
\\
&
=
\arg\min_{\mathbf{C}}\|\mathbf{W}_{1}(\mathbf{Y}-\mathbf{D}\mathbf{C})\mathbf{W}_{2}\|_{F}^{2}+\|\mathbf{W}_{3}^{-1}\mathbf{C}\|_{1},
\end{split}
\end{equation}
where
\vspace{-3mm}
\begin{equation}
\vspace{-3mm}
\begin{split}
\label{e9}
\mathbf{W}_{1}
&
=
\text{diag}(\sigma_{r}^{-1/2}\mathbf{I}_{p^2},\sigma_{g}^{-1/2}\mathbf{I}_{p^2},\sigma_{b}^{-1/2}\mathbf{I}_{p^2})
,
\\
\mathbf{W}_{2}
&
=
\text{diag}(\sigma_{1}^{-1/2},...,\sigma_{M}^{-1/2})
,
\mathbf{W}_{3} 
= 
\mathbf{S}
,
\end{split}
\end{equation}
and $\mathbf{I}_{p^2}$ is the $p^{2}$ dimensional identity matrix.\ Note that the diagonal elements of $\mathbf{W}_{1}$ and $\mathbf{W}_{2}$ are determined by the noise standard deviations in the corresponding channels and patches, respectively.\ The stronger the noise in a channel and a patch, the less that channel and patch will contribute to the denoised output.

\vspace{-2mm}
\subsection{Model Optimization}
\vspace{-2mm}

Letting $\mathbf{C}^{*}=\mathbf{W}_{3}^{-1}\mathbf{C}$, we can transfer the weight matrix $\mathbf{W}_{3}$ into the data-fidelity term of (\ref{e4}).\ Thus, the TWSC scheme (\ref{e4}) is reformulated as
\vspace{-3mm} 
\begin{equation}
\vspace{-3mm} 
\label{e10}
\min_{\mathbf{C}^{*}}\|\mathbf{W}_{1}(\mathbf{Y}-\mathbf{D}\mathbf{W}_{3}\mathbf{C}^{*})\mathbf{W}_{2}\|_{F}^{2}
+
\|\mathbf{C}^{*}\|_{1}.
\end{equation}
To make the notation simple, we remove the superscript $*$ in $\mathbf{C}^{*}$ and still use $\mathbf{C}$ in the following development.\ We employ the variable splitting method \cite{Eckstein1992} to solve the problem (\ref{e10}).\ By introducing an augmented variable $\mathbf{Z}$, the problem (\ref{e10}) is reformulated as a linear equality-constrained problem with two variables $\mathbf{C}$ and $\mathbf{Z}$:
\vspace{-3mm} 
\begin{equation}
\vspace{-3mm}
\label{e11}
\min_{\mathbf{C},\mathbf{Z}}\|\mathbf{W}_{1}(\mathbf{Y}-\mathbf{D}\mathbf{W}_{3}\mathbf{C})\mathbf{W}_{2}\|_{F}^{2}
+
\|\mathbf{Z}\|_{1}
\ 
\text{s.t.}
\ 
\mathbf{C}=\mathbf{Z}.
\end{equation}

Since the objective function is separable w.r.t. the two variables, the problem (\ref{e11}) can be solved under the alternating direction method of multipliers (ADMM) \cite{admm} framework.\ The augmented Lagrangian function of (\ref{e11}) is:
\vspace{-3mm} 
\begin{equation}
\vspace{-3mm}
\label{e12}
\begin{split}
\hspace{-2mm}
\mathcal{L}
&
(\mathbf{C},\mathbf{Z},\mathbf{\Delta},\rho)
\hspace{-1mm}
=
\hspace{-1mm}
\|\mathbf{W}_{1}(\mathbf{Y}-\mathbf{D}\mathbf{W}_{3}\mathbf{C})\mathbf{W}_{2}\|_{F}^{2}
\hspace{-1mm}
+
\hspace{-1mm}
\|\mathbf{Z}\|_{1}
\hspace{-1mm}
+
\hspace{-1mm}
\langle
\mathbf{\Delta},\mathbf{C}-\mathbf{Z}
\rangle
\hspace{-1mm}
+
\hspace{-1mm}
\frac{\rho}{2}
\|\mathbf{C}-\mathbf{Z}\|_{F}^{2},
\end{split}
\end{equation}
where $\mathbf{\Delta}$ is the augmented Lagrangian multiplier and $\rho>0$ is the penalty parameter.\ We initialize the matrix variables $\mathbf{C}_{0}$, $\mathbf{Z}_{0}$, and $\mathbf{\Delta}_{0}$ to be comfortable zero matrices and $\rho_{0}>0$.\ Denote by ($\mathbf{C}_{k}, \mathbf{Z}_{k}$) and $\mathbf{\Delta}_{k}$ the optimization variables and Lagrange multiplier at iteration $k$ ($k=0,1,2,...$), respectively.\ By taking derivatives of the Lagrangian function $\mathcal{L}$ w.r.t. $\mathbf{C}$ and $\mathbf{Z}$, and setting the derivatives to be zeros, we can alternatively update the variables as follows:
\vspace{1mm}
\\
(1) \textbf{Update $\mathbf{C}$ by fixing $\mathbf{Z}$ and $\mathbf{\Delta}$}:
\vspace{-4mm}
\begin{equation}
\vspace{-3mm}
\begin{split}
\label{e13}
\mathbf{C}_{k+1}
=
\arg\min_{\mathbf{C}}
&
\|\mathbf{W}_{1}(\mathbf{Y}-\mathbf{D}\mathbf{W}_{3}\mathbf{C})\mathbf{W}_{2}\|_{F}^{2}
+
\frac{\rho_{k}}{2}\|\mathbf{C} - \mathbf{Z}_{k} + \rho_{k}^{-1}\mathbf{\Delta}_{k}||_{F}^{2}.
\end{split}
\end{equation}
This is a two-sided weighted least squares regression problem with the solution satisfying that
\vspace{-2mm}
\begin{equation}
\vspace{-3mm}
\label{e14}
\mathbf{A}\mathbf{C}_{k+1}
+
\mathbf{C}_{k+1}\mathbf{B}_{k}
=
\mathbf{E}_{k},
\end{equation}
where 
\vspace{-3mm}
\begin{equation}
\vspace{-3mm}
\begin{split}
\label{e15}
\mathbf{A}
&
=
\mathbf{W}_{3}^{\top}\mathbf{D}^{\top}\mathbf{W}_{1}^{\top}\mathbf{W}_{1}\mathbf{D}\mathbf{W}_{3}
, 
\mathbf{B}_{k}
=
\frac{\rho_{k}}{2}(\mathbf{W}_{2}\mathbf{W}_{2}^{\top})^{-1}
,
\\
\mathbf{E}_{k}
&
=
\mathbf{W}_{3}^{\top}\mathbf{D}^{\top}\mathbf{W}_{1}^{\top}\mathbf{W}_{1}\mathbf{Y}+(\frac{\rho_{k}}{2}\mathbf{Z}_{k} -\frac{1}{2}\mathbf{\Delta}_{k})(\mathbf{W}_{2}\mathbf{W}_{2}^{\top})^{-1}
.
\end{split}
\end{equation}
Eq.\ (\ref{e14}) is a standard Sylvester equation (SE) which has a unique solution if and only if $\sigma(\mathbf{A})\cap\sigma(-\mathbf{B}_{k})=\emptyset$, where $\sigma(\mathbf{F})$ denotes the spectrum, i.e., the set of eigenvalues, of the matrix $\mathbf{F}$ \cite{simoncini2016computational}. We can rewrite the SE (\ref{e14}) as 
\vspace{-3mm}
\begin{equation}
\vspace{-3mm}
\label{e16}
(\mathbf{I}_{M}\otimes\bm{A}
+
\mathbf{B}_{k}^{\top}\otimes\bm{I}_{3p^2})\text{vec}(\mathbf{C}_{k+1})
=
\text{vec}(\mathbf{E}_{k}),
\end{equation}
and the solution $\mathbf{C}_{k+1}$ (if existed) can be obtained via $\mathbf{C}_{k+1}=\text{vec}^{-1}(\text{vec}(\mathbf{C}_{k+1}))$, where $\text{vec}^{-1}(\bullet)$ is the inverse of the vec-operator $\text{vec}(\bullet)$.
Detailed theoretical analysis on the existence of the unique solution is given in Section 3.1.
\vspace{1mm}
\\
(2) \textbf{Update $\mathbf{Z}$ by fixing $\mathbf{C}$ and $\mathbf{\Delta}$}:
\vspace{-2mm}
\begin{equation}
\vspace{-1mm}
\label{e17}
\mathbf{Z}_{k+1}
=
\arg\min_{\mathbf{Z}}\frac{\rho_{k}}{2}
\|\mathbf{Z} - (\mathbf{C}_{k+1}+\rho_{k}^{-1}\mathbf{\Delta}_{k})\|_{F}^{2}
+
\|\mathbf{Z}\|_{1}.
\end{equation}
This problem has a closed-form solution as
\vspace{-2mm} 
\begin{equation}
\vspace{-2mm}
\label{e18}
\mathbf{Z}_{k+1}
=
\mathcal{S}_{\rho_{k}^{-1}}(\mathbf{C}_{k+1}+\rho_{k}^{-1}\mathbf{\Delta}_{k}),
\end{equation}
where $\mathcal{S}_{\lambda}(x) = \text{sign}(x)\cdot\max(x-\lambda, 0)$ is the soft-thresholding operator.
\vspace{1mm}
\\
(3) \textbf{Update $\mathbf{\Delta}$ by fixing $\mathbf{X}$ and $\mathbf{Z}$}:
\vspace{-2mm}
\begin{equation}
\vspace{-2mm}
\label{e19}
\mathbf{\Delta}_{k+1}
=
\mathbf{\Delta}_{k} + \rho_{k}(\mathbf{C}_{k+1}-\mathbf{Z}_{k+1}).
\end{equation}
(4) \textbf{Update $\rho$}: $\rho_{k+1}= \mu\rho_{k}$, where $\mu\ge1$.
\vspace{1mm}

The above alternative updating steps are repeated until the convergence condition is satisfied or the number of iterations exceeds a preset threshold $K_{1}$.\ The ADMM algorithm converges when $\|\mathbf{C}_{k+1}-\mathbf{Z}_{k+1}\|_{F}\le \text{Tol}$, $\|\mathbf{C}_{k+1}-\mathbf{C}_{k}\|_{F}\le \text{Tol}$, and $\|\mathbf{Z}_{k+1}-\mathbf{Z}_{k}\|_{F}\le \text{Tol}$ are simultaneously satisfied, where $\text{Tol}>0$ is a small tolerance number.\ We summarize the updating procedures in Algorithm 1.

\begin{table}[t!]
\vspace{-1mm}
\centering
\begin{tabular}{l}
\Xhline{1pt}
\textbf{Algorithm 1}: Solve the TWSC Model (\ref{e4}) via ADMM
\\
\hline
\textbf{Input:} $\mathbf{Y},\mathbf{W}_{1},\mathbf{W}_{2},\mathbf{W}_{3}$, $\mu$, $\text{Tol}$, $K_{1}$;
\\
\textbf{Initialization:} $\mathbf{C}_{0}=\mathbf{Z}_{0}=\mathbf{\Delta}_{0}=\mathbf{0}$, $\rho_{0}>0$,  $k=0$, \text{T} = \text{False}; 
\\
\textbf{While} (\text{T} == \text{false}) \textbf{do}
\\
1. Update $\mathbf{C}_{k+1}$ by solving Eq.\ (\ref{e13});
\\
2. Update $\mathbf{Z}_{k+1}$ by soft thresholding (\ref{e18});
\\
3. Update $\mathbf{\Delta}_{k+1}$ by Eq.\ (\ref{e19});
\\
4. Update $\rho_{k+1}$ by $\rho_{k+1}=\mu\rho_{k}$, where $\mu\ge1$;
\\
5. $k \leftarrow k + 1$;
\\
\quad \textbf{if} (Converged) or ($k\ge K_{1}$)
\\
6.\quad\quad \text{T} $\leftarrow$ \text{True};
\\
\quad \textbf{end if}
\\
\textbf{end while}
\\
\textbf{Output:} Matrices $\mathbf{C}$ and $\mathbf{Z}$.
\\
\Xhline{1pt}
\end{tabular}
\vspace{-2mm}
\end{table}

\begin{figure}[t!]
\centering
\vspace{-46mm}
\hspace{-0mm}
\raisebox{-0.15cm}{\includegraphics[width=0.7\textwidth]{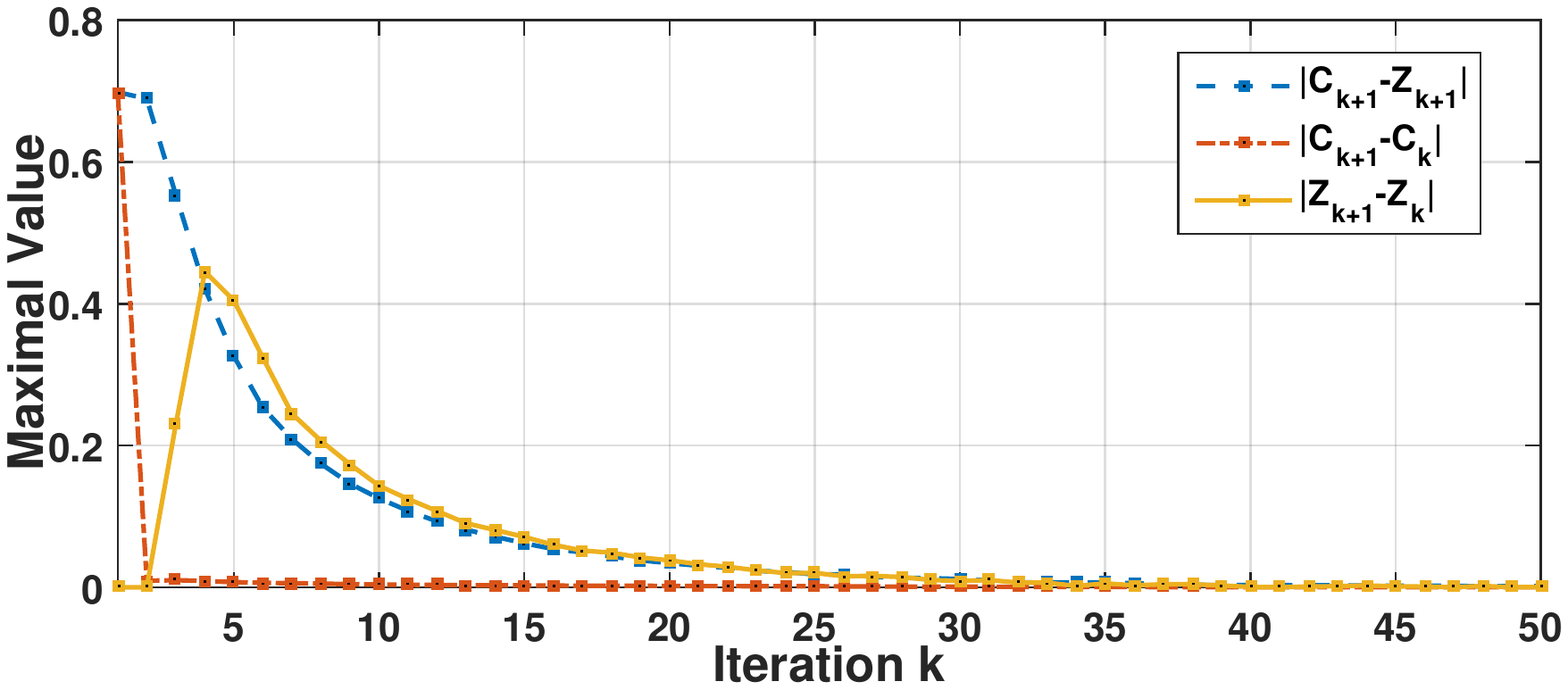}}
\vspace{-48mm}
\caption{The convergence curves of maximal values in entries of $|\mathbf{C}_{k+1}-\mathbf{Z}_{k+1}|$ (blue line), $|\mathbf{C}_{k+1}-\mathbf{C}_{k}|$ (red line), and $|\mathbf{Z}_{k+1}-\mathbf{Z}_{k}|$ (yellow line). The test image is the image in Fig.\  \ref{f2} (a).}
\label{f3}
\vspace{-3mm}
\end{figure}

\textbf{Convergence Analysis}.\ The convergence of Algorithm 1 can be guaranteed since the overall objective function (\ref{e11}) is convex with a global optimal solution.\ In Fig.\  \ref{f3}, we can see that the maximal values in $|\mathbf{C}_{k+1}-\mathbf{Z}_{k+1}|$, $|\mathbf{C}_{k+1}-\mathbf{C}_{k}|$, $|\mathbf{Z}_{k+1}-\mathbf{Z}_{k}|$ approach to $0$ simultaneously in 50 iterations.

\vspace{-3mm}
\subsection{The Denoising Algorithm}
\vspace{-2mm}

Given a noisy color image, suppose that we have extracted $N$ local patches $\{\mathbf{y}_{j}\}_{j=1}^{N}$ and their similar patches.\ Then $N$ noisy patch matrices $\{\mathbf{Y}_{j}\}_{j=1}^{N}$ can be formed to estimate the clean patch matrices $\{\mathbf{X}_{j}\}_{j=1}^{N}$.\ The patches in matrices $\{\mathbf{X}_{j}\}_{j=1}^{N}$ are aggregated to form the denoised image $\hat{\mathbf{x}}_{c}$.\ To obtain better denoising results, we perform the above denoising procedures for several (e.g., $K_{2}$) iterations.\ The proposed TWSC scheme based real-world image denoising algorithm is summarized in Algorithm 2.

\begin{table}[ht!]
\vspace{-0mm}
\centering
\begin{tabular}{l}
\Xhline{1pt}
\textbf{Algorithm 2}: Image Denoising by TWSC
\\
\hline
\textbf{Input:} Noisy image $\mathbf{y}_{c}$, $\{\sigma_{r}, \sigma_{g}, \sigma_{b}\}$, $K_{2}$;
\\
\textbf{Initialization:} $\hat{\mathbf{x}}_{c}^{(0)}=\mathbf{y}_{c}$, $\mathbf{y}_{c}^{(0)}=\mathbf{y}_{c}$;
\\
\textbf{for} $k = 1:K_{2}$ \textbf{do}
\\
1. Set $\mathbf{y}_{c}^{(k)}=\hat{\mathbf{x}}_{c}^{(k-1)}$;
\\
2. Extract local patches $\{\mathbf{y}_{j}\}_{j=1}^{N}$ from $\mathbf{y}_{c}^{(k)}$;
\\
\quad\textbf{for} each patch $\mathbf{y}_{j}$ \textbf{do}
\\
3.\quad Search nonlocal similar patches $\mathbf{Y}_{j}$;
\\
4.\quad Apply the TWSC scheme (\ref{e4}) to $\mathbf{Y}_{j}$ and obtain the estimated $\mathbf{X}_{j}=\mathbf{D}\mathbf{C}$;
\\
\quad\textbf{end for}
\\
5. Aggregate $\{\mathbf{X}_{j}\}_{j=1}^{N}$ to form the image $\hat{\mathbf{x}}_{c}^{(k)}$;
\\
\textbf{end for}
\\
\textbf{Output:} Denoised image $\hat{\mathbf{x}}_{c}^{(K_{2})}$.
\\
\Xhline{1pt}
\end{tabular}  
\vspace{-5mm}  
\end{table}
\vspace{-2mm}
\section{Existence and Faster Solution of Sylvester Equation}
\vspace{-2mm}

The solution of the Sylvester equation (SE) (\ref{e14}) does not always exist, though the solution is unique if it exists.\ Besides, solving SE (\ref{e14}) is usually computationally expensive in high dimensional cases.\ In this section, we provide a sufficient condition to guarantee the existence of the solution to SE (\ref{e14}), as well as a faster solution of (\ref{e14}) to save the computational cost of Algorithms 1 and 2. 

\vspace{-4mm}
\subsection{Existence of the Unique Solution}
\vspace{-3mm}

Before we prove the existence of unique solution of SE (\ref{e14}), we first introduce the following theorem.  
\vspace{-2mm}
\begin{theorem}
\label{th1}
Assume that $\mathbf{A}\in\mathbb{R}^{3p^2\times 3p^2}$, $\mathbf{B}\in\mathbb{R}^{M\times M}$ are both symmetric and positive semi-definite matrices.\ If at least one of $\mathbf{A}, \mathbf{B}$ is positive definite, the Sylvester equation $\mathbf{A}\mathbf{C}
+
\mathbf{C}\mathbf{B}
=
\mathbf{E}$ has a unique solution for $\mathbf{C}\in \mathbb{R}^{3p^2\times M}$.
\end{theorem}
\vspace{-2mm}
The proof of Theorem \ref{th1} can be found in the supplementary file.\ Then we have the following corollary.
\vspace{-1mm}
\begin{corollary}
\vspace{-1mm}
The SE (\ref{e14}) has a unique solution.
\end{corollary}
\vspace{-5mm}
\begin{proof}
Since $\mathbf{A},\mathbf{B}_{k}$ in (\ref{e14}) are both symmetric and positive definite matrices, according to Theorem \ref{th1}, the SE (\ref{e14}) has a unique solution. 
\end{proof}

\vspace{-4mm}
\subsection{Faster Solution}
\vspace{-1mm}

The solution of the SE (\ref{e14}) is typically obtained by the Bartels-Stewart algorithm \cite{Bartels1972}.\ This algorithm firstly employs a QR factorization \cite{GolubMatrix}, implemented via Gram-Schmidt process, to decompose the matrices $\mathbf{A}$ and $\mathbf{B}_{k}$ into Schur forms, and then solves the obtained triangular system by the back-substitution method \cite{bareiss1968sylvester}.\ However, since the matrices $\mathbf{I}_{M}\otimes\bm{A}$ and $\mathbf{B}_{k}^{\top}\otimes\bm{I}_{3p^2}$ are of $3p^2M\times 3p^2M$ dimensions, it is computationally expensive ($\mathcal{O}(p^6M^3)$) to calculate their QR factorization to obtain the Schur forms.\ By exploiting the specific properties of our problem, we provide a faster while exact solution for the SE (\ref{e14}).

Since the matrices $\mathbf{A},\mathbf{B}_{k}$ in (\ref{e14}) are symmetric and positive definite, the matrix $\mathbf{A}$ can be eigen-decomposed as $\mathbf{A}=\mathbf{U}_{\mathbf{A}}\mathbf{\Sigma}_{\mathbf{A}}\mathbf{U}_{\mathbf{A}}^{\top}$, with computational cost of $\mathcal{O}(p^6)$.\ Left multiply both sides of the SE (\ref{e14}) by $\mathbf{U}_{\mathbf{A}}^{\top}$, we can get
$
\mathbf{\Sigma}_{A}\mathbf{U}_{\mathbf{A}}^{\top}\mathbf{C}_{k+1}
+
\mathbf{U}_{\mathbf{A}}^{\top}\mathbf{C}_{k+1}\mathbf{B}_{k}
=
\mathbf{U}_{\mathbf{A}}^{\top}\mathbf{E}_{k}
$.
This can be viewed as an SE w.r.t. the matrix $\mathbf{U}_{\mathbf{A}}^{\top}\mathbf{C}_{k+1}$, with a unique solution
$
\text{vec}(\mathbf{U}_{\mathbf{A}}^{\top}\mathbf{C}_{k+1})=(\mathbf{I}_{M}\otimes\bm{\Sigma}_{\mathbf{A}}
+
\mathbf{B}_{k}^{\top}\otimes\bm{I}_{3p^2})^{-1}
\text{vec}(\mathbf{U}_{\mathbf{A}}^{\top}\mathbf{E}_{k})
$.
Since the matrix $(\mathbf{I}_{M}\otimes\bm{\Sigma}_{\mathbf{A}}
+
\mathbf{B}_{k}^{\top}\otimes\bm{I}_{3p^2})$ is diagonal and positive definite, its inverse can be calculated on each diagonal element of 
$(\mathbf{I}_{M}\otimes\bm{\Sigma}_{\mathbf{A}}
+
\mathbf{B}_{k}^{\top}\otimes\bm{I}_{3p^2})$.\ The computational cost for this step is $\mathcal{O}(p^2M)$.\ Finally, the solution $\mathbf{C}_{k+1}$ can be obtained via
$
\mathbf{C}_{k+1}=\mathbf{U}_{\mathbf{A}}\text{vec}^{-1}(\text{vec}(\mathbf{U}_{\mathbf{A}}^{\top}\mathbf{C}_{k+1}))
$.
By this way, the complexity for solving the SE (\ref{e14}) is reduced from $\mathcal{O}(p^6M^3)$ to $\mathcal{O}(\max(p^6,p^2M))$, which is a huge computational saving.

\vspace{-3mm}
\section{Experiments}
\vspace{-2mm}

To validate the effectiveness of our proposed TWSC scheme, we apply it to both synthetic additive white Gaussian noise (AWGN) corrupted images and real-world noisy images captured by CCD or CMOS cameras.\ To better demonstrate the roles of the three weight matrices in our model, we compare with a baseline method, in which the weight matrices $\mathbf{W}_{1},\mathbf{W}_{2}$ are set as comfortable identity matrices, while the matrix $\mathbf{W}_{3}$ is set as in (\ref{e8}).\ We call this baseline method the \textsl{Weighted Sparse Coding} (WSC).

\vspace{-3mm}
\subsection{Experimental Settings}
\vspace{-1mm}

\textbf{Noise Level Estimation}.\ For most image denoising algorithms, the standard deviation (std) of noise should be given as a parameter.\ In this work, we provide an exploratory approach to solve this problem. Specifically, the noise std $\sigma_{c}$ of channel $c$ can be estimated by some noise estimation methods \cite{noiselevel,Chen2015ICCV,Sutournlf}.\ In Algorithm 2, the noise std for the $m$th patch of $\mathbf{Y}$ can be initialized as 
\vspace{-3mm}
\begin{equation}
\vspace{-3mm}
\label{e20}
\sigma_{m}=\sigma\triangleq \sqrt{(\sigma_{r}^{2}+\sigma_{g}^{2}+\sigma_{b}^{2})/3}
\end{equation}
and updated in the following iterations as 
\vspace{-4mm}
\begin{equation}
\vspace{-2mm}
\label{e21}
\sigma_{m}=\sqrt{\max(0,\sigma^2-\|\mathbf{y}_{m}-\mathbf{x}_{m}\|_{2}^{2})},
\end{equation}
where $\mathbf{y}_{m}$ is the $m$th column in the patch matrix $\mathbf{Y}$, and $\mathbf{x}_{m}=\mathbf{D}\mathbf{c}_{m}$ is the $m$th patch recovered in previous iteration (please refer to Section 2.4).

\textbf{Implementation Details}.\ We empirically set the parameter $\rho_{0}=0.5$ and $\mu=1.1$.\ The maximum number of iteration is set as $K_{1}=10$.\ The window size for similar patch searching is set as $60\times60$.\ For parameters $p$, $M$, $K_{2}$, we set $p=7$, $M=70$, $K_{2}=8$ for $0<\sigma\le20$; $p=8$, $M=90$, $K_{2}=12$ for $20<\sigma\le40$; $p=8$, $M=120$, $K_{2}=12$ for $40<\sigma\le60$; $p=9$, $M=140$, $K_{2}=14$ for $60<\sigma\le100$.\ All parameters are fixed in our experiments.\ We will release the code with the publication of this work.

\begin{table}[t!]
\vspace{-4mm}
\caption{Average results of PSNR(dB) and SSIM of different denoising algorithms on 20 grayscale images corrupted by AWGN noise.}
\scriptsize
\vspace{-3mm}
\label{t1}
\begin{center}
\renewcommand\arraystretch{1.2}
\begin{tabular*}{1\textwidth}{@{\extracolsep{\fill}}cccccccccc}
\Xhline{1pt}
$\sigma_{n}$
&
Metric
&
\textbf{BM3D-SAPCA}
&
\textbf{LSSC}
&
\textbf{NCSR}
&
\textbf{WNNM}
&
\textbf{TNRD}
&
\textbf{DnCNN}
&
\textbf{WSC}
&
\textbf{TWSC}
\\
\Xhline{1pt}
\multirow{2}{*}{15}
& PSNR & 32.42 & 32.27 & 32.19 & 32.43 & 32.27 & 32.59 & 32.06 & 32.34
\\
& SSIM & 0.8860 & 0.8849 & 0.8814 & 0.8841 & 0.8815 & 0.8879 & 0.8673 & 0.8846
\\
\hline
\multirow{2}{*}{25}
& PSNR & 30.02 & 29.84 & 29.76 & 30.05 & 29.87 & 30.22 & 29.57 & 29.98
\\
& SSIM & 0.8364 & 0.8329 & 0.8293 & 0.8365 & 0.8314 & 0.8415 & 0.8179 & 0.8372
\\
\hline   
\multirow{2}{*}{35}
& PSNR & 28.48 & 28.26 & 28.17 & 28.51 & 28.33 & 28.66 & 28.01 & 28.49 
\\
& SSIM & 0.7969 & 0.7908 & 0.7855 & 0.7958 & 0.7907 & 0.8021 & 0.7765 & 0.7987
\\
\hline   
\multirow{2}{*}{50}
& PSNR & 26.85 & 26.64 & 26.55 & 26.92 & 26.75 & 27.08 & 26.35 & 26.93
\\
& SSIM & 0.7481 & 0.7405 & 0.7391 & 0.7499 & 0.7415 & 0.7563 & 0.7258 & 0.7530 
\\
\hline   
\multirow{2}{*}{75}
& PSNR & 24.74 & 24.77 & 24.66 & 25.15 & 24.97 & 25.24 & 24.54 & 25.15
\\
& SSIM & 0.6649 & 0.6746 & 0.6793 & 0.6903 & 0.6801 & 0.6931 & 0.6612 & 0.6949 
\\
\Xhline{1pt}
\end{tabular*}
\vspace{-8mm}
\end{center}
\end{table}

\vspace{-3mm}
\subsection{\parbox[t]{10cm}{Results on AWGN Noise Removal}}
\vspace{-1mm}

We first compare the proposed TWSC scheme with the leading AWGN denoising methods such as BM3D-SAPCA \cite{bm3dsapca} (which usually performs better than BM3D \cite{bm3d}), LSSC \cite{lssc}, NCSR \cite{ncsr}, WNNM \cite{wnnm}, TNRD \cite{chen2015learning}, and DnCNN \cite{dncnn} on 20 grayscale images commonly used in \cite{bm3d}.\ Note that TNRD and DnCNN are both discriminative learning based methods, and we use the models trained originally by the authors.\ Each noisy image is generated by adding the AWGN noise to the clean image, while the std of the noise is set as $\sigma\in\{15,25,35,50,75\}$ in this paper.\ Note that in this experiment we set the weight matrix $\mathbf{W}_{1}=\sigma^{-1/2}\mathbf{I}_{p^2}$ since the input images are grayscale. 


The averaged PSNR and SSIM \cite{ssim} results are listed in Table \ref{t1}.\ One can see that the proposed TWSC achieves comparable performance with WNNM, TNRD and DnCNN in most cases.\ It should be noted that TNRD and DnCNN are trained on clean and synthetic noisy image pairs, while TWSC only utilizes the tested noisy image.\ Besides, one can see that the proposed TWSC works much better than the baseline method WSC, which proves that the weight matrix $\mathbf{W}_{2}$ can characterize better the noise statistics in local image patches.\ Due to limited space, we leave the visual comparisons of different methods in the supplementary file.

\vspace{-3mm}
\subsection{Results on Realistic Noise Removal}
\vspace{-1mm}


We evaluate the proposed TWSC scheme on three publicly available real-world noisy image datasets \cite{ncwebsite,crosschannel2016,dnd2017}. 

\textbf{Dataset 1} is provided in \cite{ncwebsite}, which includes around 20 real-world noisy images collected under uncontrolled environment.\ Since there is no ``ground truth'' of the noisy images, we only compare the visual quality of the denoised images by different methods. 

\textbf{Dataset 2} is provided in \cite{crosschannel2016}, which includes noisy images of 11 static scenes captured by Canon 5D Mark 3, Nikon D600, and Nikon D800 cameras.\ The real-world noisy images were collected under controlled indoor environment.\ Each scene was shot 500 times under the same camera and camera setting.\ The mean image of the 500 shots is roughly taken as the ``ground truth'', with which the PSNR and SSIM \cite{ssim} can be computed. 15 images of size $512\times512$ were cropped to evaluate different denoising methods.  Recently, some other datasets such as \cite{PolyUdataset} are also constructed by employing the strategies of this dataset.

\textbf{Dataset 3} is called the Darmstadt Noise Dataset (DND) \cite{dnd2017}, which includes 50 different pairs of images of the same scenes captured by Sony A7R, Olympus E-M10, Sony RX100 IV, and Huawei Nexus 6P.\ The real-world noisy images are collected under higher ISO values with shorter exposure time, while the ``ground truth'' images are captured under lower ISO values with adjusted longer exposure times.\ Since the captured images are of megapixel-size, the authors cropped 20 bounding boxes of $512\times512$ pixels from each image in the dataset, yielding 1000 test crops in total.\ However, the ``ground truth'' images are not open access, and we can only submit the denoising results to the authors' \href{https://noise.visinf.tu-darmstadt.de/}{Project Website} and get the PSNR and SSIM \cite{ssim} results.

\textbf{Comparison Methods}.\ We compare the proposed TWSC method with CBM3D \cite{cbm3d}, TNRD \cite{chen2015learning}, DnCNN \cite{dncnn}, the commercial software Neat Image (NI) \cite{neatimage}, the state-of-the-art real image denoising methods ``Noise Clinic'' (NC) \cite{noiseclinic}, CC \cite{crosschannel2016}, and MCWNNM \cite{mcwnnm}.\ We also compare with the baseline method WSC described in Section 4 as a baseline.\ The methods of CBM3D and DnCNN can directly deal with color images, and the input noise std is set by Eq.\ (\ref{e20}).\ For TNRD, MCWNNM, and TWSC, we use \cite{Chen2015ICCV} to estimate the noise std $\sigma_{c}$ ($c\in\{r,g,b\}$) for each channel.\ For blind mode DnCNN, we use its color version provided by the authors and there is no need to estimate the noise std.\ Since TNRD is designed for grayscale images, we applied them to each channel of real-world noisy images.\ TNRD achieves its best results when setting the noise std of the trained models at $\sigma_{c}=10$ on these datasets.\

\begin{figure*}[ht!]
\vspace{-3mm}
\centering
\begin{minipage}[t]{0.184\textwidth}
\begin{subfigure}[t]{1\linewidth}
\raisebox{-0.15cm}{\includegraphics[width=1\textwidth]{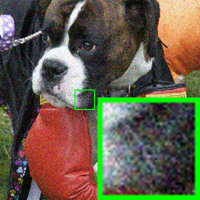}}
\centering{\footnotesize (a) Noisy  }
\end{subfigure}
\end{minipage}
\begin{minipage}[t]{0.184\textwidth}
\begin{subfigure}[t]{1\linewidth}
\raisebox{-0.15cm}{\includegraphics[width=1\textwidth]{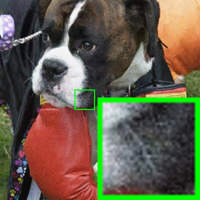}}
\centering{\footnotesize (b) CBM3D  }
\end{subfigure}
\begin{subfigure}[b]{1\linewidth}
\raisebox{-0.15cm}{\includegraphics[width=1\textwidth]{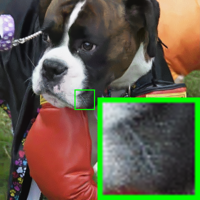}}
\centering{\footnotesize (f) NC}
\end{subfigure}
\end{minipage}
\begin{minipage}[t]{0.184\textwidth}
\begin{subfigure}[t]{1\linewidth}
\raisebox{-0.15cm}{\includegraphics[width=1\textwidth]{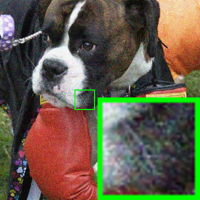}}
\centering{\footnotesize (c) TNRD  }
\end{subfigure}
\begin{subfigure}[b]{1\linewidth}
\raisebox{-0.15cm}{\includegraphics[width=1\textwidth]{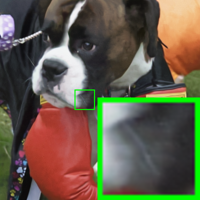}}
\centering{\footnotesize (g) MCWNNM }
\end{subfigure}
\end{minipage}
\begin{minipage}[t]{0.184\textwidth}
\begin{subfigure}[t]{1\linewidth}
\raisebox{-0.15cm}{\includegraphics[width=1\textwidth]{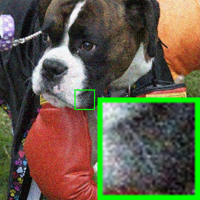}}
\centering{\footnotesize (d) DnCNN }
\end{subfigure}
\begin{subfigure}[b]{1\linewidth}
\raisebox{-0.15cm}{\includegraphics[width=1\textwidth]{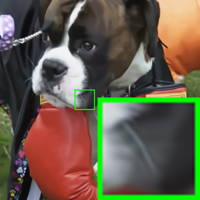}}
\centering{\footnotesize (h) WSC }
\end{subfigure}
\end{minipage}
\begin{minipage}[t]{0.184\textwidth}
\begin{subfigure}[t]{1\linewidth}
\raisebox{-0.15cm}{\includegraphics[width=1\textwidth]{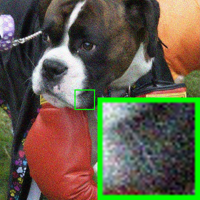}}
\centering{\footnotesize (e) NI }
\end{subfigure}
\begin{subfigure}[b]{1\linewidth}
\raisebox{-0.15cm}{\includegraphics[width=1\textwidth]{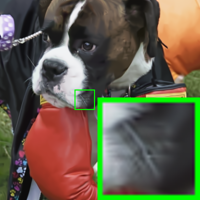}}
\centering{\footnotesize (i) TWSC }
\end{subfigure}
\end{minipage}
\vspace{-6mm}
\caption{\linespread{1}\selectfont{Denoised images of the real noisy image \textsl{Dog} \cite{ncwebsite} by different methods. Note that the ground-truth clean image of the noisy input is not available.}}
\vspace{-6mm}
\label{f4}
\end{figure*}

\hspace{-5mm}\textbf{Results on Dataset 1}.\ Fig.\ \ref{f4} shows the denoised images of ``Dog'' (the method CC \cite{crosschannel2016} is not compared since its testing code is not available).\ One can see that CBM3D, TNRD, DnCNN, NI and NC generate some noise-caused color artifacts across the whole image, while MCWNNM and WSC tend to over-smooth a little the image.\ The proposed TWSC removes more clearly the noise without over-smoothing much the image details.\ These results demonstrate that the methods designed for AWGN are not effective for realistic noise removal.\ Though NC and NI methods are specifically developed for real-world noisy images, their performance is not satisfactory.\ In comparison, the proposed TWSC works much better in removing the noise while maintaining the details (see the zoom-in window in ``Dog'') than the other competing methods.\ More visual comparisons can be found in the supplementary file.

\begin{figure*}\vspace{-1mm}
\centering
\begin{minipage}[t]{0.184\textwidth}
\begin{subfigure}[t]{1\linewidth}
\raisebox{-0.15cm}{\includegraphics[width=1\textwidth]{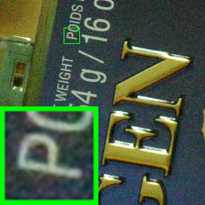}}
\centering{\scriptsize (a) Noisy 29.63dB/0.7107}
\end{subfigure}
\begin{subfigure}[b]{1\linewidth}
\raisebox{-0.15cm}{\includegraphics[width=1\textwidth]{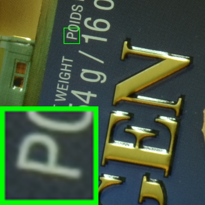}}
\centering{\scriptsize (f) Ground Truth}
\end{subfigure}
\end{minipage}
\begin{minipage}[t]{0.184\textwidth}
\begin{subfigure}[t]{1\linewidth}
\raisebox{-0.15cm}{\includegraphics[width=1\textwidth]{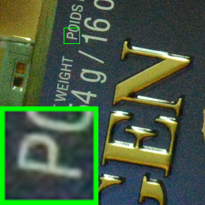}}
\centering{\scriptsize (b) CBM3D 31.12dB/0.7948}
\end{subfigure}
\begin{subfigure}[b]{1\linewidth}
\raisebox{-0.15cm}{\includegraphics[width=1\textwidth]{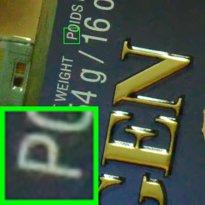}}
\centering{\scriptsize (g) NC 33.49dB/0.9024}
\end{subfigure}
\end{minipage}
\begin{minipage}[t]{0.184\textwidth}
\begin{subfigure}[t]{1\linewidth}
\raisebox{-0.15cm}{\includegraphics[width=1\textwidth]{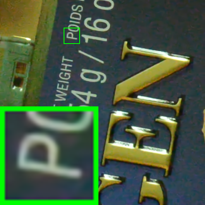}}
\centering{\scriptsize (c) TNRD 32.80dB/0.8959}
\end{subfigure}
\begin{subfigure}[b]{1\linewidth}
\raisebox{-0.15cm}{\includegraphics[width=1\textwidth]{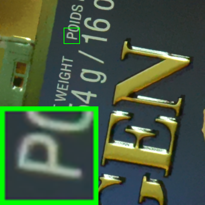}}
\centering{\scriptsize (h) CC 34.61dB/0.9206}
\end{subfigure}
\end{minipage}
\begin{minipage}[t]{0.184\textwidth}
\begin{subfigure}[t]{1\linewidth}
\raisebox{-0.15cm}{\includegraphics[width=1\textwidth]{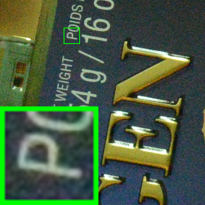}}
\centering{\scriptsize (d) DnCNN 29.83dB/0.7204}
\end{subfigure}
\begin{subfigure}[b]{1\linewidth}
\raisebox{-0.15cm}{\includegraphics[width=1\textwidth]{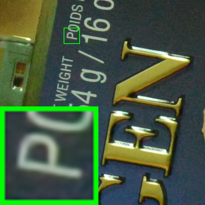}}
\centering{\scriptsize (i) MCWNNM 34.80dB/0.9217}
\end{subfigure}
\end{minipage}
\begin{minipage}[t]{0.184\textwidth}
\begin{subfigure}[t]{1\linewidth}
\raisebox{-0.15cm}{\includegraphics[width=1\textwidth]{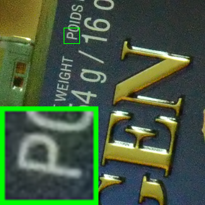}}
\centering{\scriptsize (e) NI 31.28dB/0.7781}
\end{subfigure}
\begin{subfigure}[b]{1\linewidth}
\raisebox{-0.15cm}{\includegraphics[width=1\textwidth]{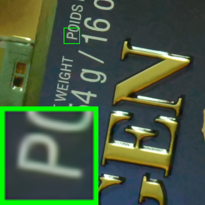}}
\centering{\scriptsize (j) TWSC \textbf{35.47}dB/\textbf{0.9369}}
\end{subfigure}
\end{minipage}
\vspace{-3mm}
\caption{\linespread{1}\selectfont{Denoised images of the real noisy image \textsl{Nikon D800 ISO 6400 1} \cite{crosschannel2016} by different methods. This scene was shot 500 times under the same camera and camera setting. The mean image of the 500 shots is roughly taken as the ``Ground Truth''.}}
\vspace{-5mm}
\label{f5}
\end{figure*}

\hspace{-5mm}\textbf{Results on Dataset 2}.\ The average PSNR and SSIM results on the 15 cropped images by competing methods are listed in Table \ref{t2}.\ One can see that the proposed TWSC is much better than other competing methods, including the baseline method WSC and the recently proposed CC, MCWNNM.\ Fig.\ \ref{f5} shows the denoised images of a scene captured by Nikon D800 at ISO = 6400.\ One can see that the proposed TWSC method results in not only higher PSNR and SSIM measures, but also much better visual quality than other methods.\ Due to limited space, we do not show the results of baseline method WSC in visual quality comparison.\ More visual comparisons can be found in the supplementary file.

\begin{table}[ht!]
\vspace{-8mm}
\caption{Average results of PSNR(dB) and SSIM of different denoising methods on 15 cropped real-world noisy images used in \cite{crosschannel2016}.}
\label{t2}
\vspace{-3mm}
\begin{center}
\renewcommand\arraystretch{1}
\small
\begin{tabular}{cccccccccc}
\Xhline{1pt}
&
\textbf{CBM3D}
&
\textbf{TNRD}
&
\textbf{DnCNN}
&
\textbf{NI}
&
\textbf{NC}
&
\textbf{CC}
&
\textbf{MCWNNM}
&
\textbf{WSC}
&
\textbf{TWSC}
\\
\hline
PSNR & 35.19 & 36.61 & 33.86 & 35.49 & 36.43 & 36.88 & 37.70 & 37.36 & \textbf{37.81} 
\\
SSIM & 0.8580 & 0.9463 & 0.8635 & 0.9126 & 0.9364 & 0.9481 & 0.9542 & 0.9516 & \textbf{0.9586} 
\\
\Xhline{1pt}
\end{tabular}
\end{center}
\vspace{-8mm}
\end{table}

\begin{figure*}[ht!]
\centering
\begin{minipage}[t]{0.184\textwidth}
\begin{subfigure}[t]{1\linewidth}
\raisebox{-0.15cm}{\includegraphics[width=1\textwidth]{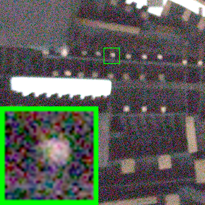}}
\centering{\footnotesize (a) Noisy}
\end{subfigure}
\end{minipage}
\begin{minipage}[t]{0.184\textwidth}
\begin{subfigure}[t]{1\linewidth}
\raisebox{-0.15cm}{\includegraphics[width=1\textwidth]{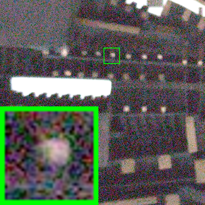}}
\centering{\footnotesize (b) CBM3D}
\end{subfigure}
\begin{subfigure}[b]{1\linewidth}
\raisebox{-0.15cm}{\includegraphics[width=1\textwidth]{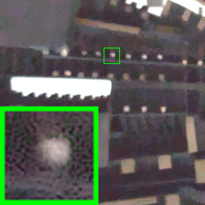}}
\centering{\footnotesize (f) NC}
\end{subfigure}
\end{minipage}
\begin{minipage}[t]{0.184\textwidth}
\begin{subfigure}[t]{1\linewidth}
\raisebox{-0.15cm}{\includegraphics[width=1\textwidth]{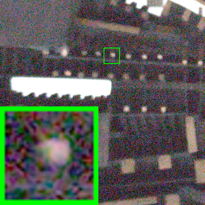}}
\centering{\footnotesize (c) TNRD}
\end{subfigure}
\begin{subfigure}[b]{1\linewidth}
\raisebox{-0.15cm}{\includegraphics[width=1\textwidth]{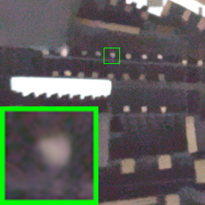}}
\centering{\footnotesize (g) MCWNNM}
\end{subfigure}
\end{minipage}
\begin{minipage}[t]{0.184\textwidth}
\begin{subfigure}[t]{1\linewidth}
\raisebox{-0.15cm}{\includegraphics[width=1\textwidth]{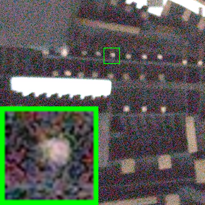}}
\centering{\footnotesize (d) DnCNN}
\end{subfigure}
\begin{subfigure}[b]{1\linewidth}
\raisebox{-0.15cm}{\includegraphics[width=1\textwidth]{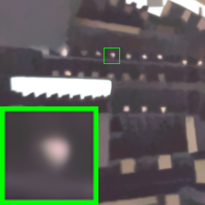}}
\centering{\footnotesize (h) WSC}
\end{subfigure}
\end{minipage}
\begin{minipage}[t]{0.184\textwidth}
\begin{subfigure}[t]{1\linewidth}
\raisebox{-0.15cm}{\includegraphics[width=1\textwidth]{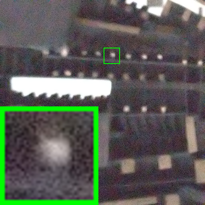}}
\centering{\footnotesize (e) NI}
\end{subfigure}
\begin{subfigure}[b]{1\linewidth}
\raisebox{-0.15cm}{\includegraphics[width=1\textwidth]{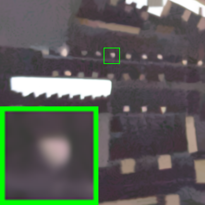}}
\centering{\footnotesize (i) TWSC}
\end{subfigure}
\end{minipage}
\vspace{-3mm}
\caption{\linespread{1}\selectfont{Denoised images of the real noisy image ``\textsl{0001\_2}'' captured by Nexus 6P \cite{dnd2017} by different methods. Note that the ground-truth clean image of the noisy input is not publicly released yet.}}
\vspace{-8mm}
\label{f6}
\end{figure*}

\hspace{-5mm}\textbf{Results on Dataset 3}.\ In Table \ref{t3}, we list the average PSNR and SSIM results of the competing methods on the 1000 cropped images in the DND dataset \cite{dnd2017}.\ We can see again that the proposed TWSC achieves much better performance than the other competing methods.\ Note that the ``ground truth'' images of this dataset have not been published, but one can submit the denoised images to the project website and get the PSNR and SSIM results.\ More results can be found in the website of the DND dataset (\url{https://noise.visinf.tu-darmstadt.de/benchmark/#results_srgb}).\ Fig.\ \ref{f6} shows the denoised images of a scene captured by a Nexus 6P camera.\ One can still see that the proposed TWSC method results better visual quality than the other denoising methods.\ More visual comparisons can be found in the supplementary file.

\begin{table}[hbpt]
\vspace{-6mm}
\caption{Average results of PSNR(dB) and SSIM of different denoising methods on 1000 cropped real-world noisy images in \cite{dnd2017}.}
\label{t3}
\vspace{-3mm}
\begin{center}
\renewcommand\arraystretch{1}
\small
\begin{tabular}{ccccccccc}
\Xhline{1pt}
&
\textbf{CBM3D}
&
\textbf{TNRD}
&
\textbf{DnCNN}
&
\textbf{NI}
&
\textbf{NC}
&
\textbf{MCWNNM}
&
\textbf{WSC}
&
\textbf{TWSC}
\\
\hline
PSNR & 32.14 & 34.15 & 32.41 & 35.11 & 36.07 & 37.38 & 36.81 & \textbf{37.94}
\\
SSIM & 0.7773 & 0.8271 & 0.7897 & 0.8778 & 0.9013 & 0.9294 & 0.9165 & \textbf{0.9403}
\\
\Xhline{1pt}
\end{tabular}
\end{center}
\vspace{-8mm}
\end{table}

\hspace{-5mm}\textbf{Comparison on Speed}.\ We compare the average computational time (second) of different methods (except CC) to process one $512\times512$ image on the DND Dataset \cite{dnd2017}.\ The results are shown in Table \ref{t4}.\ All experiments are run under the Matlab2014b environment on a machine with Intel(R) Core(TM) i7-5930K CPU of 3.5GHz and 32GB RAM.\ The fastest speed is highlighted in bold. One can see that Neat Image (NI) is the fastest and it spends about 1.1 second to process an image, while the proposed TWSC needs about 195 seconds.\ Noted that Neat Image is a highly-optimized software with parallelization, CBM3D, TNRD, and NC are implemented with compiled C++ mex-function and with parallelization, while DnCNN, MCWNNM, and the proposed WSC and TWSC are implemented purely in Matlab.

\begin{table}[t!]
\vspace{-3mm}
\caption{Average computational time (s) of different methods to process a $512\times512$ image in the DND dataset \cite{dnd2017}.}
\label{t4}
\vspace{-3mm}
\begin{center}
\renewcommand\arraystretch{1}
\small
\begin{tabular}{ccccccccc}
\Xhline{1pt}
&
\textbf{CBM3D}
&
\textbf{TNRD}
&
\textbf{DnCNN}
&
\textbf{NI}
&
\textbf{NC}
&
\textbf{MCWNNM}
&
\textbf{WSC}
&
\textbf{TWSC}
\\
\hline
Time & 6.9 & 5.2 & 79.5 & \textbf{1.1} & 15.6 & 208.1 & 188.6 & 195.2
\\
\Xhline{1pt}
\end{tabular}
\end{center}
\vspace{-8mm}
\end{table}

\vspace{-2mm}
\subsection{Visualization of The Weight Matrices}
\begin{figure}[ht!]
\vspace{-6mm}
\centering
\begin{subfigure}[t]{0.45\textwidth}
\centering
\hspace{-3mm}
\raisebox{-0.1cm}{\includegraphics[width=1\textwidth]{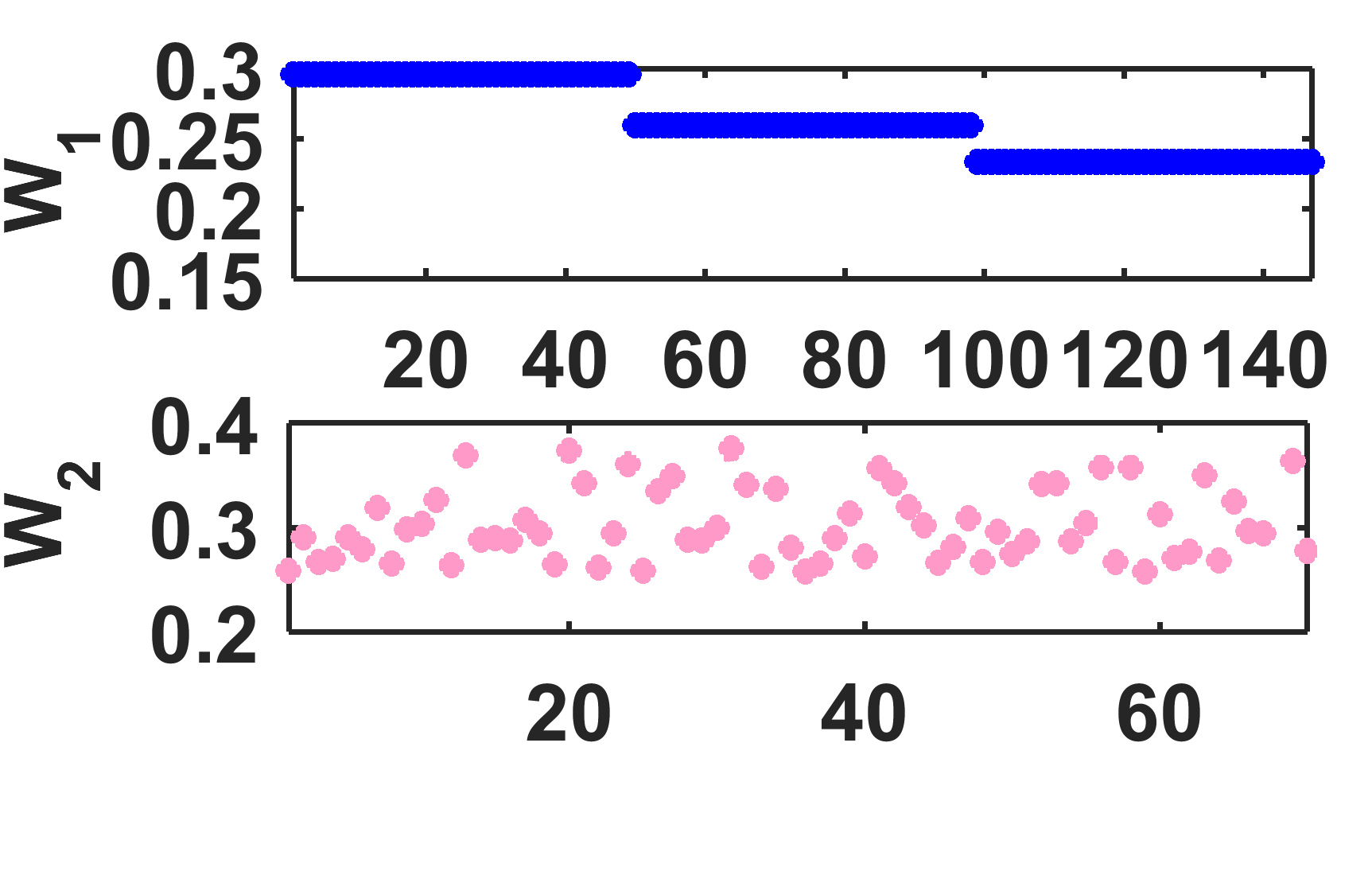}}
\end{subfigure}
\begin{subfigure}[t]{0.45\textwidth}
\centering
\hspace{-3mm}
\raisebox{-0.1cm}{\includegraphics[width=1\textwidth]{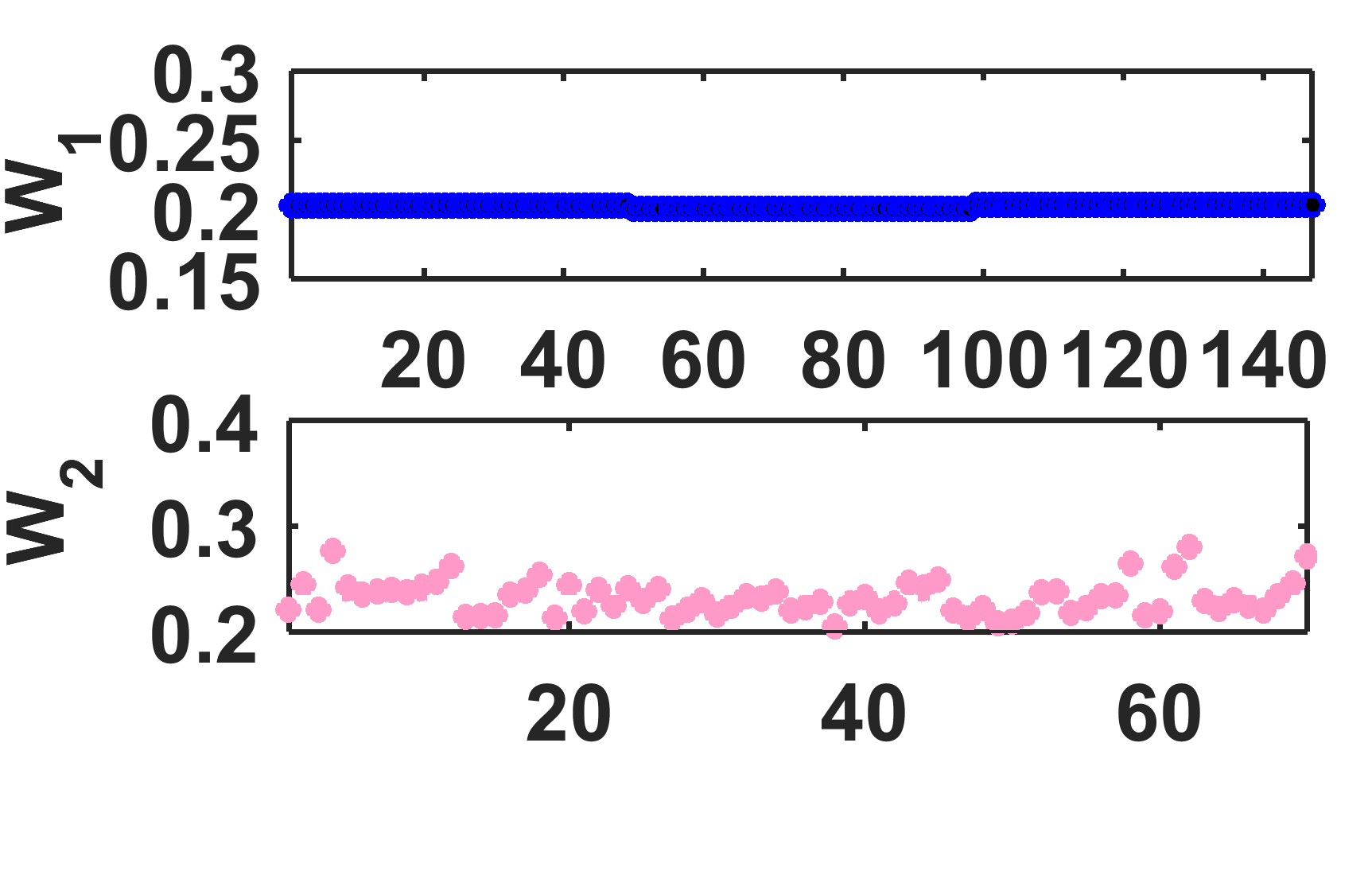}}
\end{subfigure}
\vspace{-7mm}
\caption{\small Visualization of weight matrices $\bm{W}_{1}$ and $\bm{W}_{2}$ on the real-world noisy image (left) and the synthetic noisy image (right) shown in Fig. \ref{f1}.}
\vspace{-10mm}
\label{f7}
\end{figure}

\vspace{-1mm}
The three diagonal weight matrices in the proposed TWSC model (\ref{e4}) have clear physical meanings, and it is interesting to analyze how the matrices actually relate to the input image by visualizing the resulting matrices.\ To this end, we applied TWSC to the real-world (estimated noise stds of R/G/B: 11.4/14.8/18.4) and synthetic AWGN (std of all channels:\ 25) noisy images shown in Fig.\ \ref{f1}.\ The final diagonal weight matrices for two typical patch matrices ($\bm{Y}$) from the two images are visualized in Fig.\ \ref{f7}.\ One can see that the matrix $\bm{W}_{1}$ reflects well the noise levels in the images.\ Though matrix $\bm{W}_{2}$ is initialized as an identity matrix, it is changed in iterations since noise in different patches are removed differently.\ For real-world noisy images, the noise levels of different patches in $\bm{Y}$ are different, hence the elements of $\bm{W}_{2}$ vary a lot.\ In contrast, the noise levels of patches in the synthetic noisy image are similar, thus the elements of $\bm{W}_{2}$ are similar.\ The weight matrix $\bm{W}_{3}$ is basically determined by the patch structure but not noise, and we do not plot it here. 

\vspace{-2mm}
\section{Conclusion}
\vspace{-2mm}

The realistic noise in real-world noisy images captured by CCD or CMOS cameras is very complex due to the various factors in digital camera pipelines, making the real-world image denoising problem much more challenging than additive white Gaussian noise removal.\ We proposed a novel trilateral weighted sparse coding (TWSC) scheme to exploit the noise properties across different channels and local patches.\ Specifically, we introduced two weight matrices into the data-fidelity term of the traditional sparse coding model to adaptively characterize the noise statistics in each patch of each channel, and another weight matrix into the regularization term to better exploit sparsity priors of natural images.\ The proposed TWSC scheme was solved under the ADMM framework and the solution to the Sylvester equation is guaranteed.\ Experiments demonstrated the superior performance of TWSC over existing state-of-the-art denoising methods, including those methods designed for realistic noise in real-world noisy images.

\clearpage

\bibliographystyle{splncs}
\bibliography{egbib}

\end{document}